\documentclass[runningheads]{llncs}

\usepackage{graphicx}

\usepackage{array}
\newcolumntype{P}[1]{>{\centering\arraybackslash}p{#1}}

\begin{document}
	
	\title{Novel Pipeline for Diagnosing Acute Lymphoblastic Leukemia Sensitive to Related Biomarkers}

	\author{Amirhossein Askari Farsangi\inst{1} \and
		Ali Sharifi Zarchi\inst{1} \and
		Mohammad Hossein Rohban\inst{1}}
	
	\authorrunning{A. Askari Farsangi et al.}
	
	\institute{Sharif University of Technology, Iran 
		\\
		\email{{a.h.askarifarsangi@gmail.com}\\\{asharifi,rohban\}@sharif.edu}}
	
	\maketitle


	\begin{abstract}
		
		Acute Lymphoblastic Leukemia (ALL) is one of the most common types of childhood blood cancer. The quick start of the treatment process is critical to saving the patient's life, and for this reason, early diagnosis of this disease is essential. Examining the blood smear images of these patients is one of the methods used by expert doctors to diagnose this disease.
		Deep learning-based methods have numerous applications in medical fields, as they have significantly advanced in recent years. ALL diagnosis is not an exception in this field, and several machine learning-based methods for this problem have been proposed.
		In previous methods, high diagnostic accuracy was reported, but our work showed that this alone is not sufficient, as it can lead to models taking shortcuts and not making meaningful decisions. This issue arises due to the small size of medical training datasets. To address this, we constrained our model to follow a pipeline inspired by experts' work. We also demonstrated that, since a judgement based on only one image is insufficient, redefining the problem as a multiple-instance learning problem is necessary for achieving a practical result. Our model is the first to provide a solution to this problem in a multiple-instance learning setup.
		We introduced a novel pipeline for diagnosing ALL that approximates the process used by hematologists, is sensitive to disease biomarkers,  and achieves an accuracy of 96.15\%, an F1-score of 94.24\%, a sensitivity of 97.56\%, and a specificity of 90.91\% on ALL IDB 1. Our method was further evaluated on an out-of-distribution dataset, which posed a challenging test and had acceptable performance. Notably, our model was trained on a relatively small dataset, highlighting the potential for our approach to be applied to other medical datasets with limited data availability.
		
		\keywords{Cancer Detection  \and LSTM CNN architecture \and Histopathology Images \and Biomarkers \and  Multiple-Instance learning.}
	\end{abstract}


\section{Introduction}

Leukemia is a type of cancer that affects the body's blood-forming tissues, including the bone marrow and lymphatic system~\cite{apostoaei2010review,patel2015automated}. Based on whether the leukemia is acute or chronic and whether it is lymphoid or myeloid, four main types of leukemia can be considered: ALL, AML, CLL, and CML~\cite{leukemia_subtypes}.

Diagnosing leukemia through the examination of blood smear images is a common method used by hematologists~\cite{bhattacharjee2015robust}. While additional tests may be necessary for a more complete understanding of the patient's condition, experts are able to determine the presence and type of leukemia based on the number and shape of different types of white blood cells~\cite{rehman2018classification}. This suggests that deep learning has great potential for developing computer models for diagnosing leukemia from related microscopic images.

Among all the four types of leukemia, acute lymphoblastic leukemia (ALL) has special diagnostic significance because an early start to treatment can save a patient’s life. This significance grows when we consider that 75 percent of cases involve children under the age of 14~\cite{acute_lymphoblastic_leukemia}.
The detection of blast cells in the blood and bone marrow makes it a suitable target for diagnosis using microscopic images. Therefore, studying the diagnosis of ALL using deep learning models is of particular importance in order to improve the accuracy and speed of diagnosis for this common and serious disease.

The performance of deep learning models is highly dependent on the size of the dataset used for training. In medical applications, obtaining large datasets can be a challenge, and many datasets are small in size. A popular ALL dataset is the ALL IDB provided by Scotti et al.~\cite{Scotti}, which contains images of both ALL and normal patients. It is quite small in size. Another dataset, the Raabin dataset~\cite{Raabin}, was recently introduced and contains a variety of data classes, but it has not yet been much explored.

Several classifiers have been proposed for diagnosing leukemia from related microscopic images~\cite{das2022systematic}. These classifiers can be categorized based on their target classes. Some classifiers have been designed to classify more than two classes, and they often incorporate the ALL IDB dataset as part of their training due to its importance for ALL diagnosis and its availability~\cite{Ahmed,Claro,bibi2020iomt}. However, it's important to consider the issue of dataset bias~\cite{dataset_bias} when combining datasets, but some methods may not have paid enough attention to this point. In contrast, other methods have focused on the two-class problem, specifically classifying ALL from the Normal class, and the ALL IDB dataset has been widely used for this purpose~\cite{Shafique,genovese2021acute,rawat2017classification,das2021detection,das2022lightweight,das2021transfer,das2021efficient,Joshi,Thanh,mishra2019texture,moshavash2018automatic,khandekar2021automated,mishra2017gray,mishra2018glrlm,das2020detection}.

In the following, we focus only on the two-class classifiers that distinguish ALL from normal. We can categorize these classifiers based on their input type. Some of these classifiers perform on single-cell images such as those in the ALL IDB 2 dataset, which means they can only perform on processed data in the form of cropped images~\cite{Shafique,genovese2021acute,rawat2017classification,das2021detection,das2022lightweight,das2021transfer,das2021efficient}.
On the other hand, other classifiers accept images similar to what can be seen under microscopes, such as those in ALL IDB 1~\cite{das2022lightweight,das2021transfer,das2021efficient,Joshi,Thanh,mishra2019texture,moshavash2018automatic,khandekar2021automated,mishra2017gray,mishra2018glrlm,das2020detection}.

In order to become practical models, classifiers of the first type only judge the image of single cells, while what is available are microscopic images containing a large number of white blood cells. Therefore, not only are object detection methods required in these cases, but it is also necessary to aggregate the results of all cells in a way to make a judgment about the patient. It seems that the second category of models is in better condition. These models do not need object detection methods, but the second problem somehow still exists for them. There is a possibility that in a patient with ALL, there are no signs of the disease in a single microscopic image and that different parts of the patient's blood sample must be examined to make an accurate diagnosis. This is exactly what expert doctors do in this situation. In other words, labels are weak in this case, and a multiple-instance learning setup is needed. Therefore, a third category of models that handle multiple images of the same patient should be considered, and our model belongs to this category of models.
Although there are some examples of multiple-instance learning methods for other problems related to diagnosis from blood microscopic images, there are only a few methods in the literature for diagnosing leukemia. Therefore, our work aims to fill this gap and explore the potential of this approach for improving leukemia diagnosis results~\cite{sadafi2023pixel,sadafi2020attention,cooke2022multiple,gao2023childhood}.

Finally, one of our model's most important strengths is its reliability and sensitivity to related biomarkers, and we achieved these properties by applying a special training method. We demonstrated that removing blast cells from microscopic images of patients with ALL made the model have difficulty diagnosing the patient as having ALL, whereas removing normal cells made the model accurately diagnose the patient as having ALL.
Since our model evaluates patients based on multiple images, testing this property required a completely independent dataset, as ALL IDB's samples are single images. We utilized a subset of Raabin's dataset as an out-of-distribution test set. In general, testing deep learning models on out-of-distribution test sets is a challenging task, and most methods tend to fail during these tests~\cite{ood_test_set}. However, our model achieved acceptable accuracy in this challenging setting.

\section{Training Considerations for Small Medical Datasets}

In medical applications of machine learning, in addition to common evaluation metrics such as accuracy and precision, a potential qualitative criterion for assessing model reliability is the similarity between the model's decision-making process and that of a human expert. Evaluating models based on this criterion can provide insight into their performance and trustworthiness.

As an illustration of this approach, we conducted a reimplementation of the method proposed by Ahmed et al.~\cite{Ahmed} using their dataset and visualized the model's attention map with the GradCAM algorithm~\cite{GradCAM}. Through our analysis, we identified that the model makes decisions based on unexpected patterns in the input image, which we refer to as "shortcuts," and that these shortcuts do not have any significant medical meaning. This highlights a potential flaw in these models, specifically, the issue of overfitting.

Overfitting can occur for a variety of reasons, including model complexity and dataset quality. Complex models are required for extracting high-level features from image data for proper processing; however, when the dataset is small in comparison to the model's complexity, the model's variance increases, resulting in overfitting. Another factor that can contribute to overfitting is when the training data is not clean, causing the model to learn shortcuts. To avoid this issue, it is crucial to use a clean dataset that minimizes the chances of spurious correlations. To address the potential causes of overfitting, two potential solutions are to increase the dataset size through augmentation methods and to manually clean the dataset.

It should be noted that we have an implicit assumption that we aim to train a classifier that performs similarly to the human expert classifier. This means that, in the expert's opinion, applied augmentation should not change the label of the image. In our case, cell morphology is an important criterion for classifying a specific cell as normal or diseased. As a result, we are not allowed to use augmentations like shearing that change the cell's morphology, and we have to use augmentations like rotating and translating instead. However, we can see that the majority of the proposed methods have not paid enough attention to this point.

\begin{figure}[t]
	\center{
		\includegraphics[width=150pt]{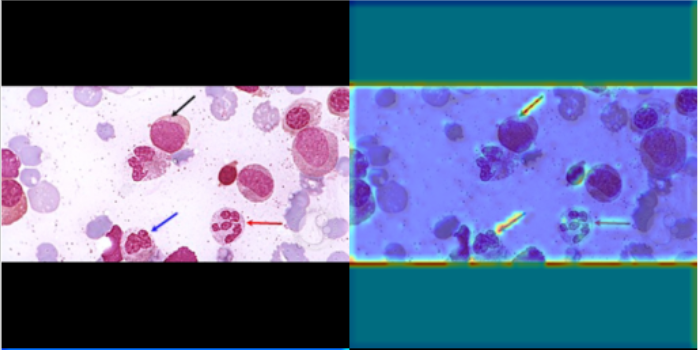}
		\includegraphics[width=150pt]{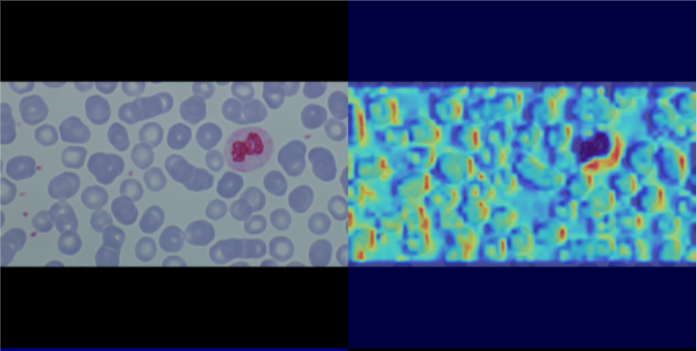}
	}
	\caption{Visualization of model weaknesses. The attention map shows the model focusing on cell boundaries and artifacts, rather than the nucleus, indicating potential limitations in the model's ability to accurately identify relevant features.} \label{models_weaknesses}
\end{figure}

Using these considerations, we discovered that the issues with shortcuts persisted and that we needed to find another solution. Two of these visualizations are shown in Fig.~\ref{models_weaknesses}. In the following, we express our method to overcome this problem.

\section{Method}
\subsection{Pipeline}

We presented a pipeline that performs the decision-making process step by step. This approach allows us to observe the model's procedure for producing the final result and helps us design each step based on the actual process used by hematologists.

Since the presence of blast cells plays a decisive role in the existence of this disease, specialists look for these cells among white blood cells when examining the patient's microscopic image and making a decision based on that.

\begin{figure}[t]
	\center{\includegraphics[width=0.7\textwidth]{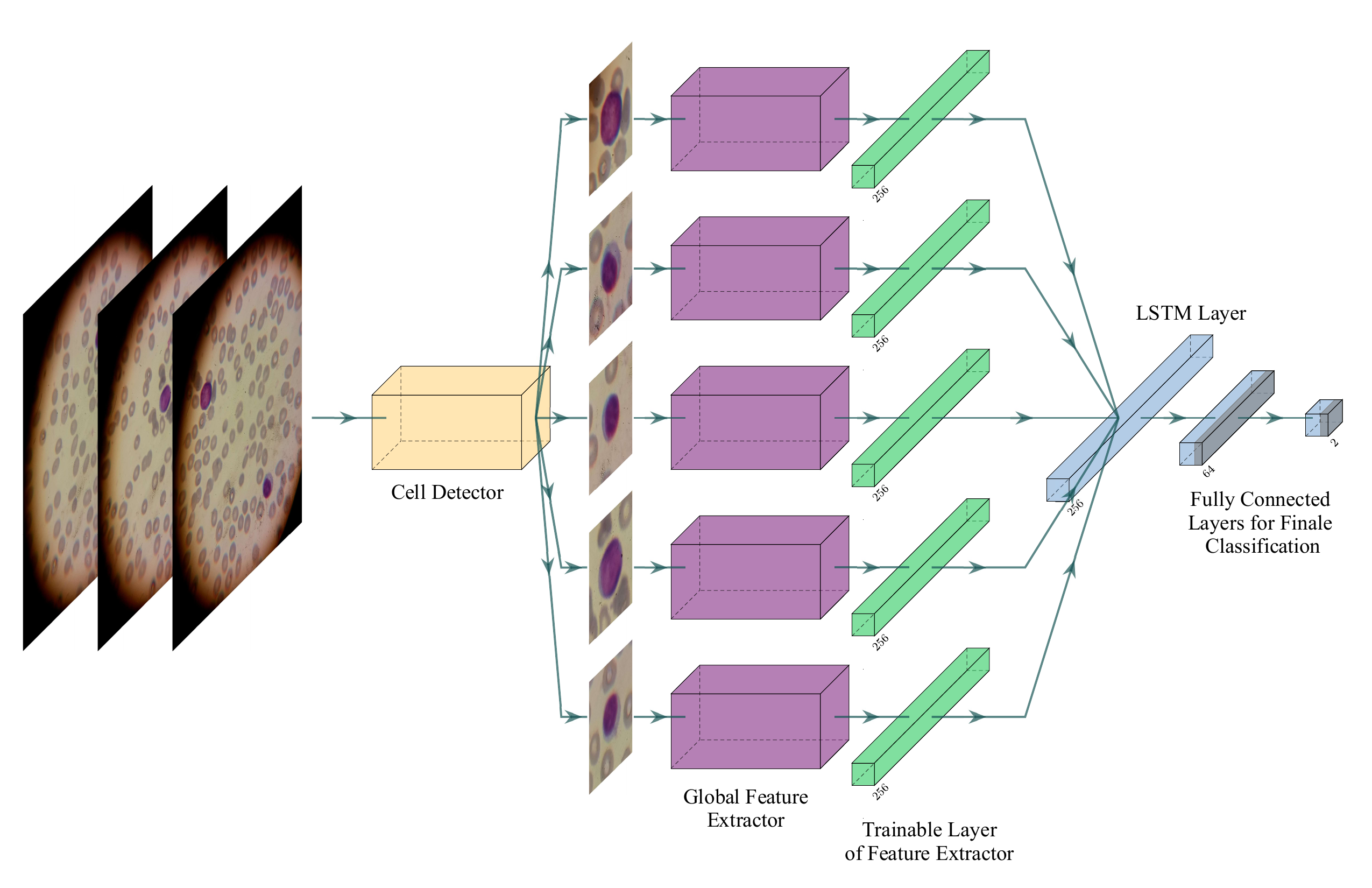}}
	\caption{Proposed pipeline for automatic diagnosis of acute lymphoblastic leukemia (ALL) using microscopic images. The pipeline consists of four main steps: (1) Object Detection to detect white blood cells in the input image, (2) Feature Extraction (3) Profiling using an LSTM, and (4) Final Classification. This pipeline enables us to mimic the decision-making process used by hematologists and provides a reliable and accurate way to diagnose ALL using microscopic images.} \label{Architecture}
\end{figure}

The first step in the computer simulation of this process should be a cell detector that detects white blood cells in the input image. The second step is to analyze each cell image using the criterion that is sensitive to whether a cell is a blast or not. In the third step, we must summarize the previous step's results and describe the patient's condition using a set of parameters. Finally, based on the generated report, we must determine the existence of this disease. Fig.~\ref{Architecture} depicts the overall layout of this pipeline. The following goes over each step in detail.


\subsubsection{Object Detection} For the white blood cell detector, we used a pre-trained Faster RCNN network with a ResNet50 backbone~\cite{ren2015faster}. We fine-tuned it using the ALL IDB 1 dataset, which was manually annotated.

\subsubsection{Feature Extraction}

To extract informative features from images, large networks with numerous training parameters are often required. However, training these networks from scratch on small datasets often leads to overfitting. A common solution is to use pre-trained networks on ImageNet, which are global feature extractors that produce rich feature vectors for each input image. In this study, we utilized a pre-trained AlexNet network with fixed weights as the global feature extractor~\cite{krizhevsky2017imagenet}. To customize these feature vectors for our problem, we added a trainable 256-node, fully connected layer that takes these feature vectors as input.
In our study, we refrained from fine-tuning the global feature extractor since the computationally intensive process of training the LSTM-based architectures does not permit simultaneous optimization of all weights.

\subsubsection{Profiling} 

There are various ways to aggregate the extracted features from the cell images of a patient. In this work, we used an LSTM-based architecture for this purpose. As the LSTM network accepts each series length as its input, this model doesn't have a problem with the different numbers of white blood cell images from different patients. In addition, it allows us to analyze the set of microscopic images for each patient, increasing the model's reliability and accuracy. It is reasonable because hematologists do not just look at one part of a patient's blood sample; instead, they move the blood slide under the microscope and make decisions based on what they see at several points. We used the LSTM network with 256 internal nodes, which produces a vector of length 64 as a patient's feature vector.

\subsubsection{Final Classification} To classify the patient's feature vector, we used only one fully connected layer with two nodes in the final classification.

\subsection{Dataset}
Our research utilized two different datasets: the ALL IDB and the Raabin dataset. The ALL IDB dataset is the most widely used dataset in the literature, consisting of two subsets. The first subset, ALL IDB 1, contains 108 normal or diseased blood microscopic images, with blast cell centers annotated in ALL cases. The second subset, ALL IDB 2, includes 260 images of single white blood cells labeled as normal or cancerous. The Raabin dataset, on the other hand, comprises 938 single-cell images of normal white blood cells.

\subsubsection{Cell Detection Dataset} 
To train the faster RCNN network, a dataset with blood microscopic images containing bounding boxes around white blood cells is required. Although ALL IDB 1 is a suitable option, it only provides annotations for blast cells, not normal ones. Therefore, we manually annotated the normal cells in these images to utilize this dataset for our purpose.

\subsubsection{Generated Dataset For training LSTM} Inputs in the form of series of the same size are required for training LSTM networks. Furthermore, because LSTM networks are data-hungry models, a large dataset of image series of the same length is required for successful convergence. Because there is no such dataset in our case, we decided to create one. We generate a dataset based on the following assumption:

\noindent
\textit{Assumption:} The series of white blood cell images belong to ALL class if and only if there were at least one blast cell in it.

\begin{figure}[t]
	\center{\includegraphics[width=\textwidth]{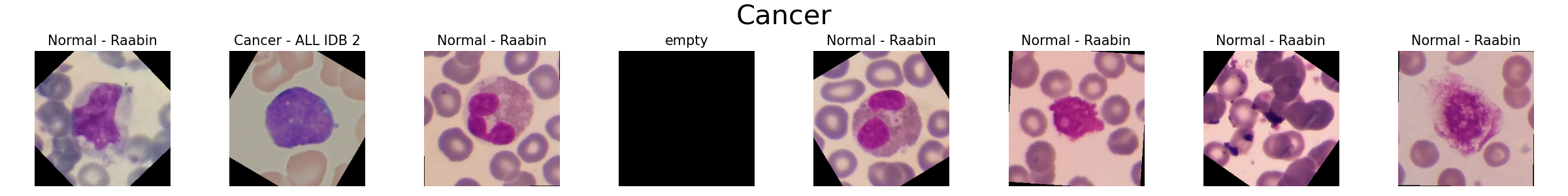}}
	\caption{Example of an image series from the generated dataset for LSTM network training. This sample is from the cancer class and contains blank, blast cell, and normal cell images with augmentations} \label{sample_generated_dataset}
\end{figure}

Based on this assumption, we generate different training image series of the same length by randomly selecting proper cell images from ALL IDB 2 and single cell images from the Raabin dataset. We choose a different number of cells because we expect the number of cells in each input image to be different. To make the sequences the same size, we add the required number of empty images to the set of selected images.
Because LSTM networks are sensitive to the order of inputs, and in our case, the order of images is unimportant, we decided to shuffle the image series to reduce LSTM sensitivity to the image series order.
Fig.~\ref{sample_generated_dataset} depicts an example of one of these image series. 

\subsection{Training}
The proposed pipeline's training was completed in three stages. Faster RCNN was trained on its dataset to become a white blood cell detector in the first step. LSTM and the classifier had been trained in two other steps. These two steps are explained below.

\subsubsection{Making sensitivity to the blast cells}
To train the network to pay attention to blast inputs, we first trained the LSTM and classifier on a one-length image series. Because of our dataset generation assumption, the classifier output should be considered cancer if and only if the input image is a blast cell. If this training process is successful, we expect the LSTM and classifier to extract information from AlexNet features that correlates with whether the input cell is a blast or not.

\subsubsection{Training for analyzing image series}
We optimize our network on the generated image series of length 15 in the final training step. Our goal in this step is to teach the model to apply what it learned about blast cells in the previous stage to the analysis of a series of images.


\section{Results}

\subsection{Classification Performance}

\begin{figure}
	\center{\includegraphics[width=180pt]{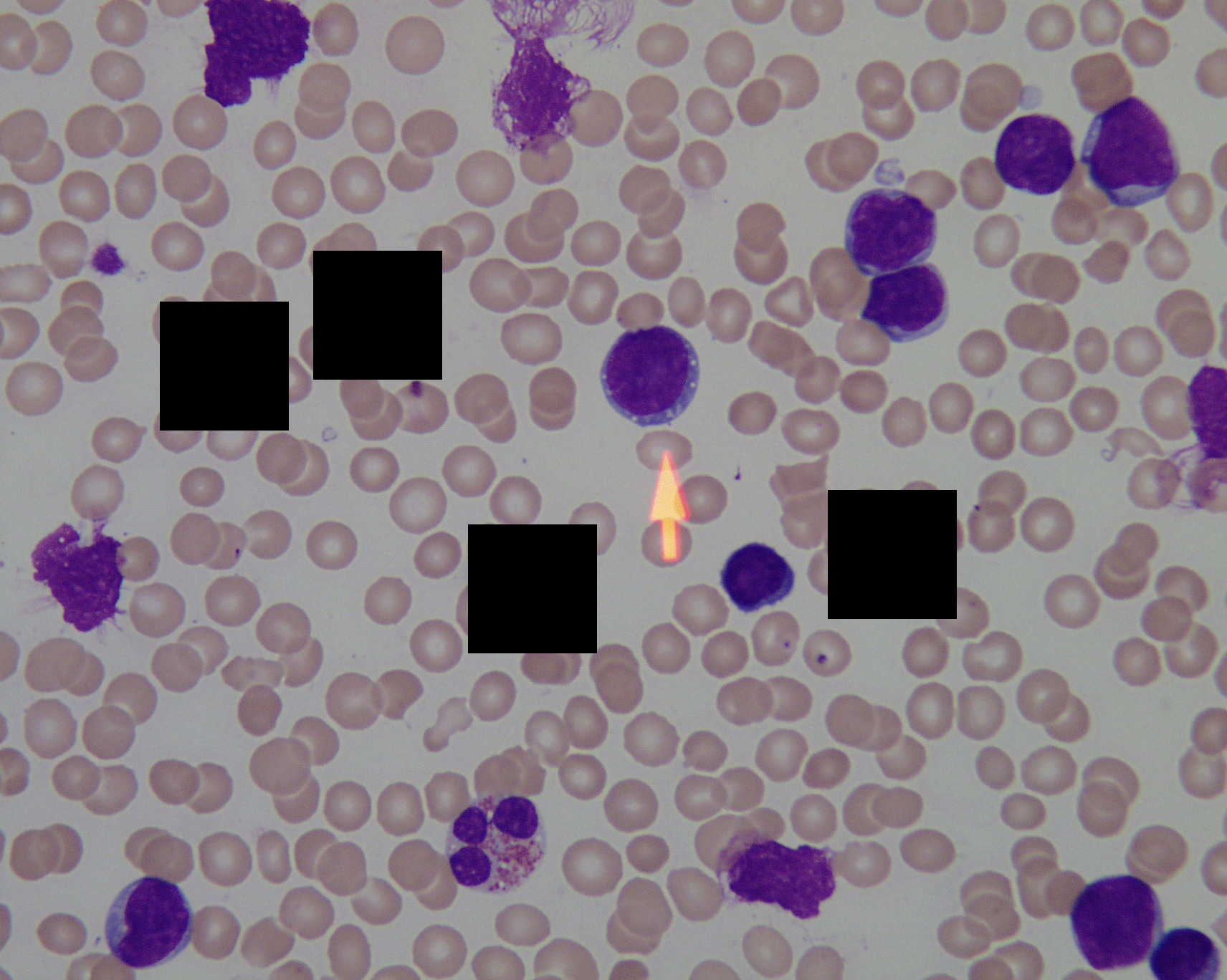}}
	\caption{Sample of the modified ALL IDB 1 dataset used for model evaluation. Black squares indicate the removed cells from the ALL IDB 2 dataset used in training to ensure a valid evaluation. Furthermore, we ensured that each image labeled as cancer contained at least one blast cell.} \label{ALL_IDB_1_modified_dataset}
\end{figure}

To ensure a valid evaluation of our model on the ALL IDB 1 dataset, which includes cropped cells from the ALL IDB 2 dataset used in training, we first removed the corresponding cells from the evaluation set. Fig.~\ref{ALL_IDB_1_modified_dataset} provides a sample of this modified dataset. Our model achieved accuracy and F1-score of 96.15\% and 94.24\%, respectively. Table~\ref{results} shows the accuracy achieved by our method in comparison to other methods.

In the following, we evaluated our model on an out-of-distribution test set consisting of multiple images for each patient, requiring classification in a multiple-instance setup. To do this, we utilized a subset of the Raabin Leukemia dataset, which contains numerous microscopic images for each patient. Since the number of patients is small in this dataset, we decided to split all images of each patient and considered them as separate patients. We did this in a way that the total white blood cells of different patients are equal and called this constant number partition size.

On a partition size of 50, our model attained an accuracy of 72.88\% and an F1-score of 71.01\%. The average accuracy and F1-score across partition sizes ranging from 20 to 100 were found to be 71.78\% and 70.35\%, respectively. It is important to note that testing on an out-of-distribution test set is a challenging task, and it is common for model performance to decrease significantly under such conditions.

\begin{table}
	\caption{Comparison of accuracies and application domains of different methods. Although our model doesn't have the best accuracy among other models, it has the advantage of being able to accept multiple input images, which is an important feature for real-world applications.}
	\label{results}
	\begin{center}
		\begin{tabular}{ |P{3cm}|P{2cm}|P{5cm}| }
			\hline		
			Method & Accuracy & Input Type\\
			\hline
			Shafique et al.~\cite{Shafique}					& \textbf{99.5} 	& Single Cell Image \\
			Genovese et al.~\cite{genovese2021acute}		& 96.84 			& Single Cell Image \\
			Rawat et al.~\cite{rawat2017classification}		& 97.6 				& Single Cell Image \\
			Das et al.~\cite{das2021detection}				& 96.15 			& Single Cell Image \\
			Das et al.~\cite{das2022lightweight}			& 98.21 			& Single Cell Image \\
			Das et al.~\cite{das2021transfer}				& 96.67 			& Single Cell Image \\
			Das et al.~\cite{das2021efficient}				& 97.18 			& Single Cell Image \\ \hline
			Das et al.~\cite{das2022lightweight}			& 99.39 			& Single Blood Smear Image \\
			Das et al.~\cite{das2021transfer}				& 96.97 			& Single Blood Smear Image \\
			Das et al.~\cite{das2021efficient}				& 99.39 			& Single Blood Smear Image \\
			Joshi et al.~\cite{Joshi}						& 98.0 				& Single Blood Smear Image \\
			Thanh et al.~\cite{Thanh}						& 96.6 				& Single Blood Smear Image \\
			Mishra et al.~\cite{mishra2019texture}			& \textbf{99.66} 	& Single Blood Smear Image \\
			Moshavash et al.~\cite{moshavash2018automatic}	& 89.81 			& Single Blood Smear Image \\
			Khandekar et al.~\cite{khandekar2021automated}	& -	    			& Single Blood Smear Image \\
			Mishra et al.~\cite{mishra2017gray}				& 96.00 			& Single Blood Smear Image \\
			Mishra et al.~\cite{mishra2018glrlm}			& 96.97 			& Single Blood Smear Image \\
			Ahmed et al.~\cite{Ahmed}						& 88.25 			& Single Blood Smear Image \\
			Das et al.~\cite{das2020detection}				& 96.00 			& Single Blood Smear Image \\ \hline
			
			Ours											& 96.15 			& Multiple Blood Smear Images \\
			\hline

		\end{tabular}
	\end{center}
\end{table}

It is necessary to mention the outcome of the cell detector section here. The faster RCNN was trained on 85 percent of the ALL IDB 1 images and achieved a mean average precision (mAP) of 96.03\% on the remaining 15 percent of ALL IDB 1 images.

\subsection{Sensitivity to Blast Cells}

Our special training method has led us to expect that our model will be sensitive to ALL biomarkers, particularly blast cells. To put this hypothesis to the test, we designed a test that involved removing blast cells from the image series of ALL patients to see if it would make it difficult for our model to identify this new sample as belonging to the ALL class. We also expected that removing normal cells from these images would improve the model's performance. To perform this test, we required the coordinates of blast and normal cells to be annotated in each dataset. While these annotations were available for the ALL IDB 1 dataset, the Raabin leukemia dataset did not have such annotations. Therefore, we trained a Faster RCNN on ALL IDB 1 data to detect blast and normal cells. This object detector achieved a mean average precision of 93.41\% on the test set split from ALL IDB 1.

Our test on the ALL IDB 1 dataset showed that the model's recall under blast removal, normal removal, and no attack conditions was 43.90\%, 97.56\%, and 97.56\%, respectively. The observed decrease of 53.66\% in recall under blast removal indicates that our hypothesis was correct and that the model was indeed sensitive to blast cells.

For this test on the Raabin dataset, we evaluated the model's performance on groups of images of varying sizes. For each patient, we selected as many images as the group size and used object detection to form three different cell series: one with only blast cells, one with only normal cells, and one with all cells. It should be noted that the length of these three cell series for each patient is not equal, and the sum of the first two is equal to the third.

\begin{figure}[t]
	\center{\includegraphics[width=180pt]{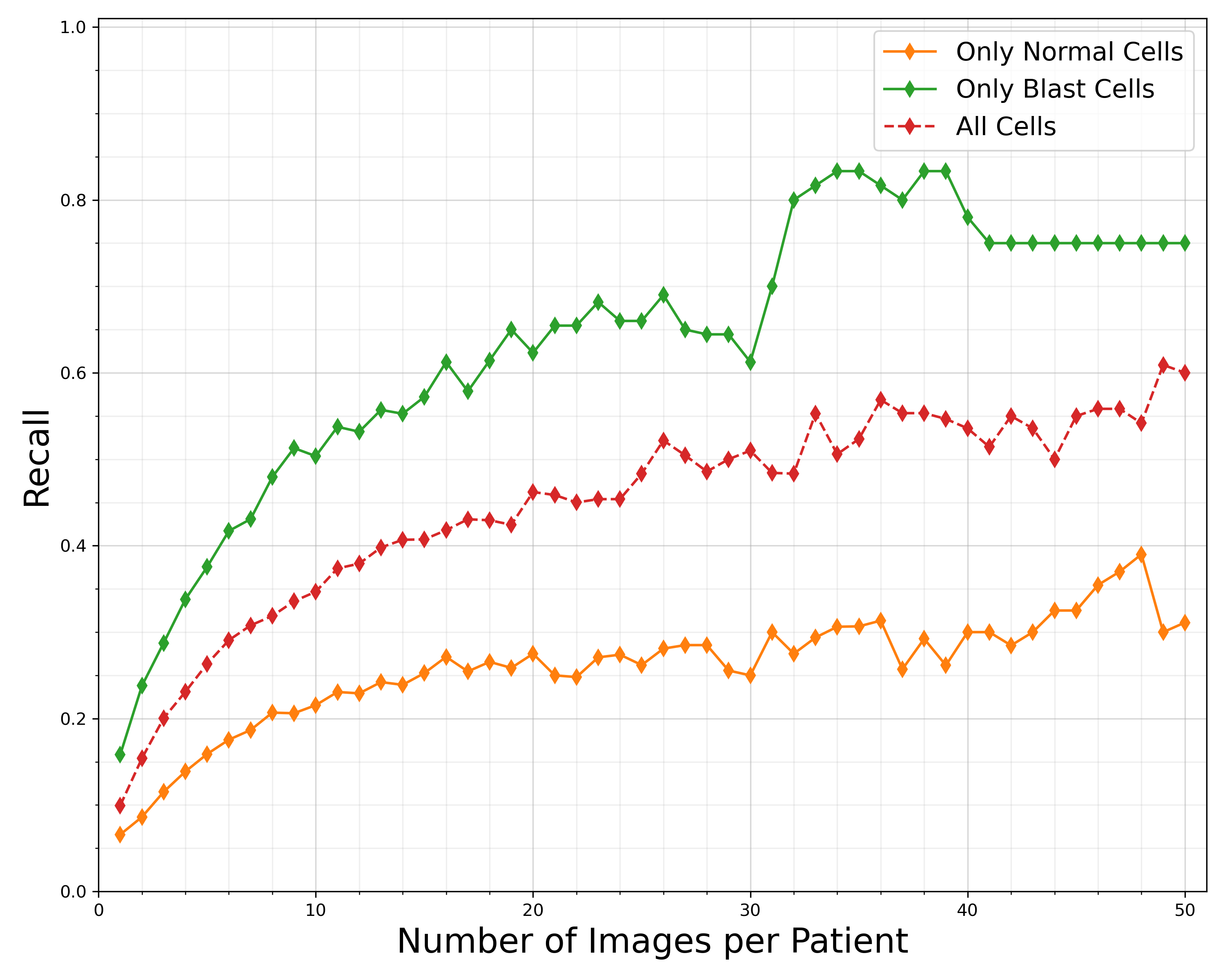}}
	\caption{Effect of removing blast cells and normal cells on the model's recall. The removal of blast cells decreases the model recall by an average of 18.84\% across all group sizes, while the removal of normal cells increases the model recall by an average of 18.69\%. The results validate our hypothesis and demonstrate the significance of blast cells in the model's decision-making process.} \label{blast_removing_attack}
\end{figure}

We plotted the recall of the model under blast removal, normal cell removal, and no attack conditions. Our results showed that removing blast cells reduced the model's recall by an average of 18.84\% across all group sizes. Conversely, removing normal cells has a large effect on model recall and, on average, increases it by 18.69\%. This finding validates our model's proposed hypothesis on an even out-of-distribution test set and demonstrates the significance of blast cells in our model's decision-making process. Fig.~\ref{blast_removing_attack} provides a visual representation of our findings.
It should be noted that while we used the results of the trained detector on the ALL IDB 1 dataset to evaluate the model's performance on the Raabin leukemia dataset, there is no ground truth for the Raabin dataset. Therefore, an error in object detection may be present in the obtained results.

\section{Ablation Study}

\subsection{Cell numbers effect}

\begin{figure}[t]
	\center{\includegraphics[width=180pt]{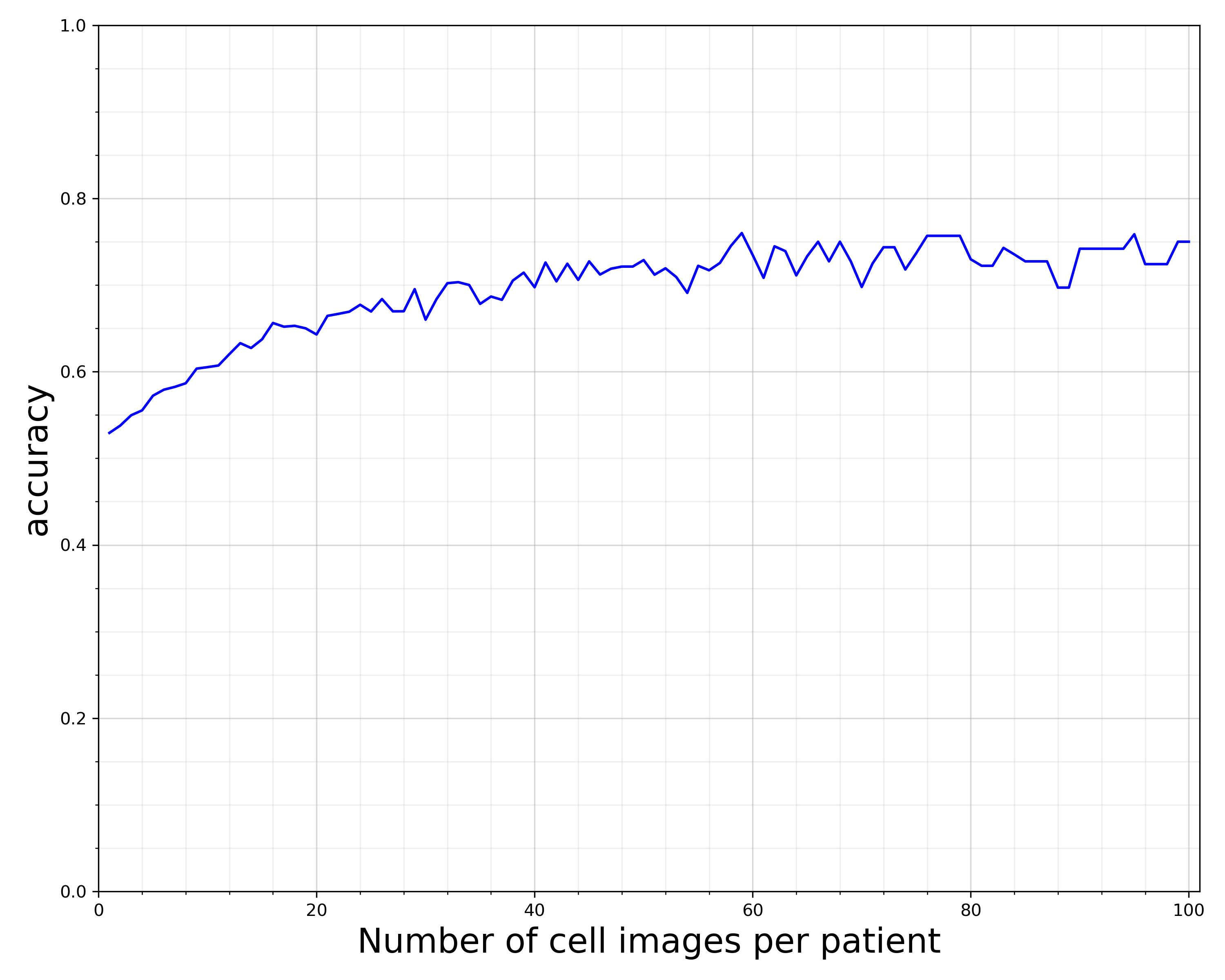}}
	\caption{The effect of the number of cell images on model performance. The plot shows the relationship between the number of white blood cell images per patient and the accuracy of the model in detecting ALL. The results demonstrate a positive correlation between the number of images and model performance.} \label{group_size_effect}
\end{figure}

Based on our understanding that blast cells are not uniformly distributed throughout a blood smear, we hypothesize that increasing the number of white blood cells per patient will improve our model's performance in detecting ALL. Fig.~\ref{group_size_effect} shows the accuracy plot for varying the number of images per patient, and as expected, we observe a positive correlation between the number of images and model performance. In all subsequent reports, the reported metrics are the averages of the results obtained for group sizes ranging from 20 to 100, since evaluations were performed on different group sizes.

\subsection{Different feature extractors}

Our method can employ several pre-trained feature extractors, including AlexNet, InceptionV3~\cite{szegedy2016rethinking}, ResNet50~\cite{he2016deep}, VGG16~\cite{simonyan2014very}, and ViT-base-patch-16~\cite{dosovitskiy2020image}. We conducted a comparison of these models, and the results are presented in Table~\ref{different_FEs}. From this table, it is evident that the AlexNet feature extractor yields the best performance. We will conduct further tests using this feature extractor in the subsequent sections.

\begin{table}
	\caption{Comparison of different feature extractors. ViT-base-patch-16 achieves the best maximum results, while AlexNet has the best average results.}
	\label{different_FEs}
	\begin{center}
		\begin{tabular}{ |c|c|c|c|c| }
			\hline		
			Feature Extractor & Mean Accuracy & Max Accuracy & Mean F1-score & Max F1-score \\
			\hline
			AlexNet 			& \textbf{71.78} 	& 76.00 			& \textbf{70.35} 	& 75.00 \\
			InceptionV3 		& 34.17 			& 48.21 			& 33.01 			& 45.83 \\
			ResNet50 			& 69.43 			& \textbf{80.77}	& 65.27 			& \textbf{79.27}\\
			VGG16 				& 58.23 			& 69.23 			& 56.86 			& 67.50 \\
			ViT-base-patch-16 	& 64.67 			& 75.00 			& 58.74 			& 70.91 \\
			\hline
		\end{tabular}
	\end{center}
\end{table}

\subsection{LSTM effect}

The main reason for using LSTM in our architecture was to add the ability to aggregate the extracted results in our model. It may seem that since in the first step of training, we trained the model on single-cell images to learn how to identify blast cells, in the second step of training the model has generalized this result with a counting approach. In other words, it would be possible that the LSTM layer may not perform more than a linear operator. How to test this hypothesis is a bit unclear, but we can still define some tests.

For each group of patient images, we fed the extracted cells individually to the model for labeling as either blast or normal cells. As a result, each patient in the test set was assigned two values, representing the number of normal and blast cells, respectively. Using this data, we trained a perceptron to classify each patient. Since the training and testing sets were identical for this perceptron, the resulting accuracy can be considered ideal. The average accuracy of the ideal perceptron was found to be 8.51 percent lower than the average accuracy of the LSTM model. Therefore, it can be concluded that the LSTM layer is capable of more than just linear operations.

\subsection{Pre-training effect}

It is necessary to investigate the potential benefits of pre-training, the first step of our training process. We hypothesize that pre-training can accelerate model convergence by providing a controlled environment in which the model can recognize the important biomarker for ALL, which is blast cells, and learn to differentiate them from normal cells. To evaluate this hypothesis, we trained two models, one with pre-training and one without. All other conditions were kept identical. Our results indicate that the model without pre-training has an average accuracy that is 2.05 percent lower than that of the pre-trained model. Therefore, pre-training appears to be a beneficial step in our training process, leading to improved model accuracy.

\subsection{Impact of training series length}

\begin{figure}
	\center{\includegraphics[width=180pt]{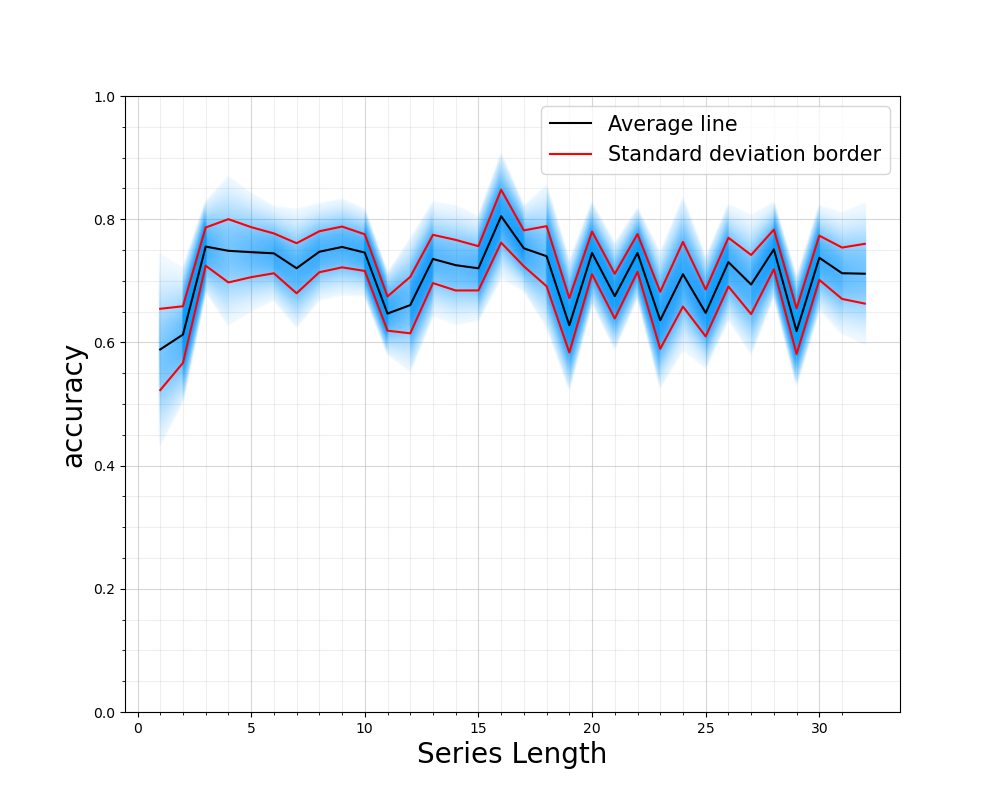}}
	\caption{Impact of training series length on model performance. The black line represents the mean accuracy for each series length, with red lines indicating the standard deviation. The blue shaded area depicts the normal distribution fitted to the accuracies.} \label{impact_of_training_series_length}
\end{figure}

To assess the impact of training series length on model performance, we trained several classifiers with varying series lengths ranging from 1 to 32. The results of these classifiers are shown in Fig.~\ref{impact_of_training_series_length}. The plot suggests that there is an initial positive correlation between series length and model performance. However, beyond a certain threshold, the impact of series length on performance diminishes, and longer series lengths do not significantly improve the model's accuracy.

\section{Conclusion}

In this work, we aimed to develop a machine learning-based model for diagnosing acute lymphoblastic leukemia from blood smear images. Since the size of the datasets is small in this field, training networks in an end-to-end manner leads the model to find shortcuts for making decisions instead of using medically meaningful patterns. To address this issue, we introduce a pipeline inspired by the hematologists' approach, consisting of four main steps: detecting white blood cells, analyzing each cell, aggregating results, and decision-making. Compared to end-to-end training, this approach has several advantages.

First and foremost, the training process is a kind of search among all feasible classifiers, and if we want to achieve a classifier similar to a human expert, we must constrain this search space. Each data point is a kind of constraint, and that's why the size of the dataset becomes important. In our problem, we do not have access to such large datasets, so we should apply the constraints in another way. We did this by constraining the classifier architecture to the described pipeline.
In addition, this approach allows us to monitor the performance of individual components and to find and fix possible faults.

Another important thing that we did for training our pipeline was to train the final classifier in two stages. The first step can be considered an auxiliary task that makes the classifier sensitive to the biomarkers of ALL, and the second step is for the model to learn to generalize the knowledge learned in the first step.

Finally, we show that our model is sensitive to ALL biomarkers. Furthermore, we analyzed the impact of our design choices, such as the use of the AlexNet feature extractor, the LSTM layer, the pre-training step, and the length of the training series.

In our work, we addressed the need to redefine the problem of acute lymphoblastic leukemia (ALL) diagnosis as a multiple-instance learning problem, which has not been done before. To overcome this, we generated a suitable training dataset and evaluated our model on an out-of-distribution test set, achieving acceptable results.
It seems that the model's sensitivity to staining is its major weakness, and further work should focus on addressing this issue to improve its performance. As a result, one potential solution could be to train feature extractors that are less sensitive to staining.

\bibliographystyle{unsrt}
\bibliography{refs}

\end{document}